\documentclass{article}

\usepackage{PRIMEarxiv}

\usepackage{enumitem}
\usepackage{amsmath}
\usepackage{mathtools}
\usepackage{amsthm}
\usepackage{longtable}
\usepackage{comment}
\usepackage{indentfirst}

\usepackage{natbib}

\usepackage[utf8]{inputenc}
\usepackage[T1]{fontenc}
\usepackage{hyperref}
\usepackage{url}
\usepackage{booktabs}
\usepackage{amsfonts}
\usepackage{amssymb}
\usepackage{nicefrac}
\usepackage{microtype}
\usepackage{graphicx}
\usepackage{multirow}
\usepackage{array}
\usepackage{siunitx}
\usepackage{xcolor}
\usepackage{listings}
\usepackage{graphics}
\usepackage{subcaption}
\usepackage{caption}

\definecolor{codegreen}{rgb}{0,0.6,0}
\definecolor{codegray}{rgb}{0.5,0.5,0.5}
\definecolor{codepurple}{rgb}{0.58,0,0.82}
\definecolor{backcolour}{rgb}{0.97,0.97,0.97}

\usepackage{amsmath} 
\usepackage{graphicx}
\usepackage{amssymb}
\usepackage{tikz}
\usetikzlibrary{positioning,arrows.meta,calc,fit}

\title{\includegraphics[width=\textwidth]{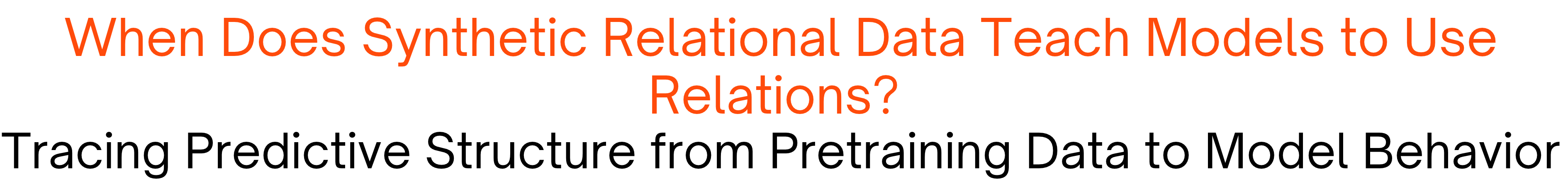}}

\author{
  Shivam Dubey$^*$, Mohamed Bouadi$^*$,\\ 
  Nassim Bouarour$^*$, Varun Kulkarni, \\
  Aditya Tanna, Vinay Kumar Sankarapu \\
  \affiliation{Lexsi Labs}\\
   \small{$^*$Equal contribution}
}

\setabstract{%
Relational foundation models are increasingly pretrained on synthetic databases, yet downstream benchmarks reveal little about \emph{why} one synthetic corpus produces a better model than another. In particular, strong performance may arise from realistic row-level statistics without the model ever learning to use relational structure. We study this as a data-attribution problem: which property of synthetic pretraining data induces relational computation? Using four Relational Transformer checkpoints trained with the same architecture, initialization, objective, and compute budget on corpora produced by four relational data generators, we trace a measurable property of the data to learned computation and downstream behavior. We hypothesize that relational mechanisms emerge when cross-table information is \emph{predictively necessary} for the masked-cell pretraining objective. RelDiff exhibits by far the largest predictive gain from foreign-key-linked parents, and its corresponding model is uniquely sensitive to foreign-key interventions on unseen databases. This dependence survives a random-initialization control, grows monotonically with the fraction of corrupted links, and localizes to a serial cross-table pathway. Finally, disrupting the same mechanism during downstream inference removes RelDiff's advantage on relational tasks while leaving structure-insensitive models nearly unchanged. These results connect a property of synthetic training data to a learned mechanism and, through intervention, to downstream behavior
}

\setkeywords{Relational Foundation Models; Synthetic Pretraining; Data Attribution; Mechanistic Interpretability; Relational Learning}

\runningtitle{When Does Synthetic Relational Data Teach Models to Use Relations? Tracing Predictive Structure from Pretraining Data to Model Behavior}

\begin{document}
\maketitle

\section{Introduction}
\label{sec:introduction}

Relational databases encode information at two levels: in the attributes of individual rows and in the relationships that connect entities across tables. A customer's properties, for example, may be informative on their own, but many prediction problems additionally depend on related transactions, products, accounts, or organizations. Recent work has therefore moved beyond single-table prediction \cite{tabpfn,tabiclv2,limix,orionmsp, orionbix} toward relational deep learning and relational foundation models (RFMs), including Relational Deep Learning \citep{fey2024position}, RelBench \citep{robinson2024relbench}, Griffin \citep{wang2025griffin}, KumoRFM \citep{fey2025kumorfm}, and the Relational Transformer (RT) \citep{ranjan2026relational}. A central goal of these models is to transfer across databases and prediction tasks while exploiting foreign-key (FK) structure, rather than requiring a separately trained model and manually engineered joins or aggregations for
every task.

Pretraining such models creates a fundamental data problem. Large and diverse collections of real relational databases are difficult to obtain, frequently proprietary, and heterogeneous in both schema and semantics. This has motivated the use of synthetic data for foundation-model pretraining. In the single-table setting, O'PRIOR
\citep{bouadi2026shaping} shows that the design of the synthetic pretraining distribution itself can be a first-order determinant of downstream accuracy and robustness, even when model architecture and training budget are held fixed. This question becomes especially important in the relational setting, where a synthetic generator must determine not only feature distributions but also schemas, cardinalities, foreign-key structure, and
statistical dependencies across related tables. RelDiff \citep{hudovernik2025reldiff} and graph-conditional relational diffusion models (GRDM) \citep{grdm} generate data conditioned on real database structure, whereas PluRel \citep{kothapalli2026plurel} and RDB-PFN \citep{rdbpfn} construct relational pretraining distributions without requiring a reference database. These approaches differ substantially in how schemas, cardinalities, attribute distributions, and dependencies between related tables are generated.

A natural way to compare such generators is to pretrain the same model on each resulting corpus and evaluate the checkpoints on downstream tasks. This establishes which synthetic pretraining distribution produces the strongest model, but it does not establish \emph{why}. A generator may improve downstream performance because it better reproduces single-column marginals, because it produces more diverse schemas, because its rows contain
more predictable within-table structure, because its foreign-key relationships carry useful signal, or because of some combination of these factors. A benchmark score alone cannot identify which property of the synthetic data produced a particular learned capability. Our central question is therefore not simply \emph{which generator performs best?}, but rather:

\begin{quote}
\emph{What property of a synthetic relational pretraining distribution causes a model to learn to use relations?}
\end{quote}

We view this as a problem of \emph{data attribution}. Classical influence functions \citep{koh2017understanding}, TracIn \citep{pruthi2020estimating}, datamodels \citep{ilyas2022datamodels}, and TRAK \citep{park2023trak} attribute predictions or model behaviors to individual training examples or subsets. Our attribution target is
different: a synthetic generator induces an entire distribution over relational databases. Attributing a capability directly to generator identity is therefore too coarse. We instead seek an intermediate property of the generated distribution that explains why a particular computation becomes useful during pretraining.

Our hypothesis is motivated by RT's masked-cell reconstruction objective. If a masked value is already predictable from information in its own row, a model can reduce pretraining loss without making meaningful use of related tables. If FK-linked rows contain additional predictive information unavailable locally, however, exploiting relational context becomes useful for solving the objective. We refer to this property as \emph{relational predictive necessity}. Importantly, this distinguishes \emph{architectural capacity} from \emph{learned reliance}: providing a model with cross-table computational pathways does not imply that pretraining will make those pathways functionally important. Section~\ref{sec:setup} formalizes this hypothesis and describes the relevant RT mechanisms.

We test this idea by tracing a sequence from the training data to model behavior. First, we measure relational predictive signal directly in the generated corpora, without using RT. We then test whether the corresponding pretrained checkpoints functionally depend on FK structure during masked reconstruction. Next, we intervene on RT's internal cross-table computation to localize the mechanism responsible for this dependence. Finally, we perturb the same relational information during downstream RelBench inference to determine whether the identified mechanism is responsible for the observed downstream behavior.

Our contributions are threefold:
\begin{enumerate}
    \item \textbf{A data-side measure of relational predictive necessity.}
    We quantify the predictive gain provided by FK-linked information beyond features available in the target row, giving a model-independent measure of whether a synthetic pretraining corpus rewards cross-table computation.

    \item \textbf{A mechanistic analysis of relational reliance.}
    Using structural corruption, a random-initialization control, graded interventions, and pathway-specific ablations, we show that the RT checkpoint pretrained on the corpus with the strongest parent-linked predictive signal develops substantially stronger functional reliance on relational structure. We further localize this dependence to a specific serial cross-table computation.

    \item \textbf{A causal connection between mechanism and downstream behavior.}
    We rerun downstream inference while disrupting FK structure or the corresponding internal computation. On a task where relational structure provides a large advantage, disrupting the identified mechanism removes that advantage, whereas checkpoints that do not rely on relational structure remain nearly unchanged. We also identify cases in which disrupting relational information improves performance, showing that learned relational priors can help or hurt depending on how well they match the downstream database.
\end{enumerate}

The causal scope of these results is important. Our inference-time interventions establish a causal link between the identified relational computation and model behavior while holding the trained checkpoint fixed. The synthetic pretraining corpora themselves, however, are not produced by independently randomizing relational predictive necessity while holding every other property constant. We therefore treat the relationship between corpus predictive structure and the learned mechanism as evidence for our hypothesis, rather than as a universal causal law. A stronger data-side causal test would selectively alter cross-table predictive dependencies while preserving other corpus statistics and retrain from the same initialization.

\section{Problem Setup and Hypothesis}
\label{sec:setup}

\subsection{Controlled Comparison of Synthetic Pretraining Distributions}
\label{sec:controlled-setting}

We study four checkpoints of the Relational Transformer (RT) \citep{ranjan2026relational}, a relational foundation model pretrained by masked-cell reconstruction on multi-table databases. Given a partially masked database, RT predicts the missing cell values from the remaining cells and the database's relational structure.

The four checkpoints use the same model architecture: 12 transformer blocks, hidden dimension 256, and approximately 19.2M parameters. They also share the same masked-value reconstruction objective, initialization, and training budget. The experimental variable is the synthetic corpus used for pretraining. Each checkpoint is trained on 1,000 generated databases produced by one of four relational data generators: RelDiff \citep{hudovernik2025reldiff}, GRDM \citep{grdm}, PluRel \citep{kothapalli2026plurel}, or RDB-PFN \citep{rdbpfn}.

These generators represent two broad approaches to relational synthesis. RelDiff and GRDM are \emph{reference-conditioned}: their generated databases are tied to structural properties of real reference databases. PluRel and RDB-PFN are \emph{reference-free}: they sample relational databases from designed synthetic priors rather than conditioning generation on a particular reference database. The resulting corpora consequently differ not only in relational dependencies, but also in properties such as feature distributions, schema structure, cardinalities, and data types.

This distinction is important for the interpretation of our experiments. We do \emph{not} treat generator identity itself as the causal variable. A statement such as "RelDiff teaches relational reasoning" would conflate all properties that differ between the generated corpora. Instead, we measure a specific property of those corpora---whether FK-linked information provides predictive signal beyond information available within a
row---and ask whether that property is associated with the relational computation learned during pretraining.

\subsection{Cross-Table Computation in RT}
\label{sec:rt-cross-table}

RT represents database cells as tokens and applies separate attention operations along different axes of the relational input. Our analysis focuses on two operations through which information can cross table boundaries.

\paragraph{Parent access (\texttt{feat}).}
The feature-attention stream normally operates over cells associated with the target row, but its attention mask additionally exposes cells from rows referenced through foreign keys. Consequently, when predicting a cell in a child row, the model can directly access attributes of the corresponding FK-linked parent row. We refer to this operation as \emph{parent access}.

\paragraph{Child aggregation (\texttt{nbr}).}
The neighbor-attention stream operates in the complementary direction. It allows a representation associated with a parent entity to aggregate information from rows that reference that entity through a foreign key. We refer to this operation as \emph{child aggregation}.

These two routes are architecturally available in every checkpoint. Thus, architecture alone cannot explain differences in relational behavior: all four models have the \emph{capacity} to exchange information across tables. The relevant question is whether their respective pretraining distributions create sufficient optimization pressure for those pathways to become functionally important.

Importantly, we do not assume a priori that parent access and child aggregation constitute independent or serial computations. Their functional interaction is determined experimentally in Sec.~\ref{sec:mechanism} through pathway-specific interventions.

\subsection{Evaluation Setting}
\label{sec:evaluation-setting}

Model-side analyses are performed on real RelBench databases \citep{robinson2024relbench,relbenchv2}: \texttt{rel-stack}, \texttt{rel-arxiv}, and \texttt{rel-ratebeer}. All checkpoint comparisons use identical preprocessed inputs and, within each experiment, identical masked cells, batches, and interventions. This paired design ensures that differences in reconstruction loss arise from the checkpoints' responses to the same inputs rather than from different sampled examples.

Our most controlled mechanistic experiments---including the random-initialization baseline, graded FK corruption, and pathway-specific ablations---use fixed \texttt{rel-stack} batches so that the interventions can be compared directly under a single evaluation protocol. We additionally report broader experiments across all three
databases in Appendix~\ref{app:reliance-full} to assess whether the observed behavior is specific to a single database.

One corpus-specific caveat requires separate treatment. In the available GRDM pretraining corpus, textual and datetime fields were converted to numerical representations rather than preserved under their original semantic types. RT contains type-specific encoders, including a dedicated text pathway, so this preprocessing difference means that the GRDM checkpoint did not receive the same type-level training signal as the other checkpoints.
We therefore retain GRDM as an informative comparison throughout the analysis but do not interpret its absolute performance ranking as an isolated consequence of its relational generation strategy. Appendix~\ref{app:grdm-types} provides the corresponding corpus audit.

\subsection{Relational Predictive Necessity}
\label{sec:predictive-necessity-hypothesis}

Our hypothesis follows directly from the masked-cell reconstruction objective. Let $y$ denote a masked target value, $x_{\mathrm{row}}$ the information available from the target row, and $x_{\mathrm{rel}}$ information available through FK-linked rows. If relational information contributes no additional predictive signal, then
\[
p(y \mid x_{\mathrm{row}},x_{\mathrm{rel}})
\approx
p(y \mid x_{\mathrm{row}}).
\]
In this regime, a model can reduce reconstruction loss primarily from within-row statistics. Cross-table pathways may exist architecturally but are not required to solve the pretraining task.

Conversely, when
\[
p(y \mid x_{\mathrm{row}},x_{\mathrm{rel}})
\neq
p(y \mid x_{\mathrm{row}}),
\]
FK-linked information contains predictive signal unavailable locally. Exploiting relational context can then reduce the pretraining loss beyond what is achievable from the target row alone. We call this condition \emph{relational predictive necessity}.

We do not attempt to estimate the conditional distributions above directly. Instead, Section~\ref{sec:corpus} operationalizes predictive necessity through out-of-sample predictive gain: we compare a predictor that uses only target-row information with the same predictor augmented by information from related tables. This measurement is
performed entirely on the synthetic corpora and does not involve RT, preventing the quantity being explained from also defining the explanatory variable.

The hypothesis generates three linked empirical predictions:

\paragraph{H1: Predictive structure in the pretraining data.}
Synthetic corpora differ in the amount of target information available through FK-linked rows beyond what can already be recovered from the target row. A corpus that creates stronger relational predictive necessity should therefore exhibit larger out-of-sample predictive gains when relational information is added.

\paragraph{H2: Learned relational reliance.}
If relational predictive necessity creates pressure to learn cross-table computation, the corresponding pretrained checkpoint should become functionally dependent on correct relational structure. Corrupting FK assignments or disabling the relevant internal pathways should therefore increase reconstruction loss substantially more for that checkpoint than for models whose pretraining objectives could be solved primarily from
row-local information.

\paragraph{H3: Relational mechanism and downstream behavior.}
If the learned cross-table computation is responsible for a downstream capability, intervening on that computation while holding the trained checkpoint fixed should produce a corresponding change in downstream performance. Conversely, checkpoints that do not functionally rely on relational structure during reconstruction should be comparatively insensitive to the same intervention.

Together, these predictions define the empirical chain studied in the remainder of the paper:
\[
\boxed{
\text{predictive structure in data}
\;\longrightarrow\;
\text{relational reliance}
\;\longrightarrow\;
\text{localized cross-table mechanism}
\;\longrightarrow\;
\text{downstream behavior}
}
\]

The arrows have different causal status. The inference-time interventions used to test H2 and H3 manipulate the trained model or its relational inputs while holding the checkpoint fixed, permitting causal conclusions about the dependence of model behavior on the intervened mechanism. By contrast, the four pretraining corpora were not generated by independently manipulating relational predictive necessity while holding all other distributional properties fixed. The data-to-mechanism link in H1--H2 is therefore a controlled observational comparison. Establishing that link causally would require retraining on corpora in which cross-table predictive coupling is selectively manipulated while other corpus properties, initialization, objective, and training budget are held constant.

\section{Experimental Evaluation}
\label{sec:experiments}

We evaluate the relational-predictive-necessity hypothesis through four increasingly model-dependent stages. First, we measure whether FK-linked information is predictively useful in the synthetic pretraining corpora themselves, without using RT. Second, we test whether the corresponding trained checkpoints functionally depend on relational structure during masked-cell reconstruction. Third, we intervene on RT's internal cross-table
pathways to localize the computation responsible for that dependence. Finally, we apply the same structural and internal interventions during downstream RelBench inference.

This progression separates three questions that are often conflated in evaluations of synthetic pretraining data: whether relations contain useful signal in the generated corpus, whether a model actually learns to use that signal, and whether the resulting computation matters for downstream prediction.

%%%%%%%%%%%%%%%%%%%%%%%%%%%%%%%%%%%%%%%%%%%%%%%%%%%%%%%%%%%%%%%%%%%%%%%%%%%%%%%
% 3.1 PREDICTIVE STRUCTURE
%%%%%%%%%%%%%%%%%%%%%%%%%%%%%%%%%%%%%%%%%%%%%%%%%%%%%%%%%%%%%%%%%%%%%%%%%%%%%%%

\subsection{Predictive Structure in Synthetic Pretraining Data}
\label{sec:corpus}

Our first experiment tests H1 directly at the data level. The goal is to determine whether a target value in each synthetic corpus becomes more predictable when information from FK-linked tables is made available, independently of any representation learned by RT.

\paragraph{Protocol.}
For each eligible target column $y$, we construct three prediction problems using the same LightGBM learner. The first uses only features available in the target row. The second augments those features with attributes from the FK-linked parent row. The third instead augments the target row with simple statistics of linked children: mean,
standard deviation, and count. Let the corresponding held-out scores be
\[
S_{\mathrm{row}}
=
R^2(y \mid x_{\mathrm{row}}),
\qquad
S_{\mathrm{parent}}
=
R^2(y \mid x_{\mathrm{row}},x_{\mathrm{parent}}),
\]
and
\[
S_{\mathrm{child}}
=
R^2\!\left(
y \mid x_{\mathrm{row}},a(x_{\mathrm{child}})
\right),
\]
where $a(\cdot)$ denotes the child-aggregation features. We define
\[
\Delta_{\mathrm{parent}}
=
S_{\mathrm{parent}}-S_{\mathrm{row}},
\qquad
\Delta_{\mathrm{child}}
=
S_{\mathrm{child}}-S_{\mathrm{row}}.
\]
Positive gain therefore means that relational information predicts the held-out target beyond what is available from the target row alone.

We sample 40 generated databases from each corpus. Individual held-out $R^2$ values are clipped to $[-1,1]$ before differences are computed, preventing a small number of pathological fits from dominating the aggregate. Table~\ref{tab:necessity} reports medians over target-column prediction problems; parent gains additionally include 95\% bootstrap confidence intervals. Full task-level distributions are shown in Appendix~\ref{app:necessity-distribution}.

\begin{table}[t]
\caption{
\textbf{Predictive structure in the synthetic pretraining corpora.} Row-only $R^2$ measures target predictability without cross-table information. $\Delta_{\mathrm{parent}}$ and $\Delta_{\mathrm{child}}$ measure the additional
predictive value of FK-linked parent attributes and simple child aggregates, respectively. $N_{\mathrm{p}}/N_{\mathrm{c}}$ gives the number of target-column problems for which each relational feature set is available.
}
\label{tab:necessity}
\centering
\small
\setlength{\tabcolsep}{4.2pt}
\begin{tabular}{lccccc}
\toprule
Corpus &
Targets &
$N_{\mathrm{p}}/N_{\mathrm{c}}$ &
Row-only $R^2$ &
$\Delta_{\mathrm{parent}}$ &
$\Delta_{\mathrm{child}}$ \\
\midrule
RelDiff
& 48
& 30/22
& 0.022
& \textbf{+0.072} {\scriptsize $[+0.018,+0.091]$}
& $-0.001$ \\

GRDM
& 120
& 65/78
& $-0.011$
& $-0.057$ {\scriptsize $[-0.076,-0.044]$}
& $-0.105$ \\

PluRel
& 113
& 84/71
& 0.166
& +0.009 {\scriptsize $[+0.001,+0.022]$}
& $-0.000$ \\

RDB-PFN
& 120
& 72/68
& \textbf{0.436}
& +0.001 {\scriptsize $[-0.004,+0.009]$}
& $-0.005$ \\
\bottomrule
\end{tabular}
\end{table}

\paragraph{RelDiff contains substantially stronger parent-linked predictive signal.}
Table~\ref{tab:necessity} reveals markedly different pretraining regimes. In RelDiff, adding linked-parent attributes improves median held-out $R^2$ by $0.072$. The corresponding gain is $0.009$ for PluRel and approximately zero for RDB-PFN. RelDiff's parent gain is therefore about eight times PluRel's and more than two orders of magnitude larger than the rounded RDB-PFN gain.

The contrast with row-local predictability is also informative. RDB-PFN has by far the largest row-only score ($R^2=0.436$) while exhibiting essentially no gain from parent information. Its generated values are therefore comparatively predictable without cross-table context. RelDiff lies at the opposite extreme: row-local predictability is weak ($R^2=0.022$), while linked parents provide substantial additional signal.

Simple child aggregates do not improve prediction in any corpus. This result narrows our hypothesis: the directly measurable relational incentive in these corpora is principally \emph{access to linked parents}, rather than the mean/count-style child aggregates tested here. It does not imply that richer functions of child rows are uninformative.

The GRDM results require the preprocessing caveat described in Section~\ref{sec:evaluation-setting}: its available corpus does not preserve the same semantic type information as the other arms. We therefore do not interpret its negative gains as evidence against relational generation in general.

\paragraph{Could this simply reflect corpus diversity?}
RelDiff is also the most diverse corpus under our feature-, table-, and cross-table diversity measurements. Diversity alone, however, does not reproduce the ordering of functional relational reliance. For example, PluRel has greater measured cross-table diversity than GRDM but exhibits little structural dependence in the experiments below. We report the complete diversity analysis in Appendix~\ref{app:diversity}. This motivates our focus on predictive necessity rather than diversity alone: the relevant question is not only how much the generated databases vary, but whether variation across relations helps solve the pretraining objective.

%%%%%%%%%%%%%%%%%%%%%%%%%%%%%%%%%%%%%%%%%%%%%%%%%%%%%%%%%%%%%%%%%%%%%%%%%%%%%%%
% 3.2 FUNCTIONAL RELIANCE
%%%%%%%%%%%%%%%%%%%%%%%%%%%%%%%%%%%%%%%%%%%%%%%%%%%%%%%%%%%%%%%%%%%%%%%%%%%%%%%

\subsection{Functional Reliance on Relational Structure}
\label{sec:reliance}

The corpus experiment establishes that relational information can be predictive, but this does not show that RT actually learned to use it. We therefore intervene directly on relational structure during masked-cell reconstruction.

\subsubsection{Structural Interventions}
\label{sec:structural-interventions}

Our primary controlled probe uses 24 fixed \texttt{rel-stack} batches, corresponding to 384 masked cells at context length 1,024. Every checkpoint is evaluated on exactly the same examples and masks under four conditions:

\begin{itemize}
    \item \textbf{Original:} ordinary inference on the unmodified database;
    \item \textbf{FK shuffle:} FK-neighbor assignments are permuted among a selected
    set of row positions while cell values are left unchanged;
    \item \textbf{FK removal:} relational indices are removed;
    \item \textbf{Neighbor ablation:} the input database is unchanged, but the output
    of RT's \texttt{nbr} stream is set to zero.
\end{itemize}

For intervention $c$, we report $\Delta\mathcal{L}_{c}=\mathcal{L}_{c}-\mathcal{L}_{0}$, where $\mathcal{L}_{0}$ is the original masked-cell reconstruction loss. Positive values therefore indicate that the perturbed information or computation was useful.

\begin{table}[t]
\caption{
\textbf{Functional reliance on relational structure on \texttt{rel-stack}.} Entries are original masked-cell loss and absolute changes under three interventions. The final column reports the relative loss increase under full FK shuffling. }
\label{tab:reliance}
\centering
\small
\setlength{\tabcolsep}{4.0pt}
\begin{tabular}{lccccc}
\toprule
Model &
$\mathcal{L}_0$ &
$\Delta\mathcal{L}_{\mathrm{nbr}}$ &
$\Delta\mathcal{L}_{\mathrm{shuffle}}$ &
$\Delta\mathcal{L}_{\mathrm{remove}}$ &
Rel.\ FK change \\
\midrule
RelDiff
& 0.241
& \textbf{+0.127}
& \textbf{+0.202}
& \textbf{+0.197}
& \textbf{+83.6\%} \\

GRDM
& 0.342
& $-0.002$
& +0.018
& +0.034
& +5.3\% \\

PluRel
& 0.161
& $-0.000$
& +0.010
& +0.019
& +6.1\% \\

RDB-PFN
& 0.208
& +0.003
& +0.007
& +0.015
& +3.4\% \\

Random init
& 0.556
& $-0.025$
& +0.027
& +0.057
& +4.9\% \\
\bottomrule
\end{tabular}
\end{table}

\paragraph{RelDiff is substantially more dependent on correct FK structure.} 
Full FK shuffling raises RelDiff's loss from $0.241$ to $0.443$, an absolute increase of $0.202$ or $83.6\%$. GRDM, PluRel, and RDB-PFN increase by only $5.3\%$, $6.1\%$, and $3.4\%$, respectively. FK removal produces the same qualitative ordering. The result is therefore not specific to one particular corruption procedure.

Internal intervention gives a complementary result. Zeroing \texttt{nbr} increases RelDiff's loss by $0.127$, whereas the same ablation has negligible effect on the other trained checkpoints. Thus, the difference is not merely that RelDiff reacts strongly to an unusual input perturbation: an internal cross-table computation is itself load-bearing.

\subsubsection{Architectural Floor and Graded Corruption}
\label{sec:graded-corruption}

FK shuffling also changes the relational masks supplied to RT, so some response can arise from the architecture even in the absence of learning. We quantify this floor using a randomly initialized RT with exactly the same architecture. Full FK shuffling increases its loss by $4.9\%$, comparable to the $3.4$--$6.1\%$ changes of the three weakly reliant trained checkpoints but far below RelDiff's $83.6\%$ increase. RelDiff's relative
response is approximately $17\times$ the random-initialization effect.

\begin{figure}[t]
    \centering
    \includegraphics[width=\linewidth]{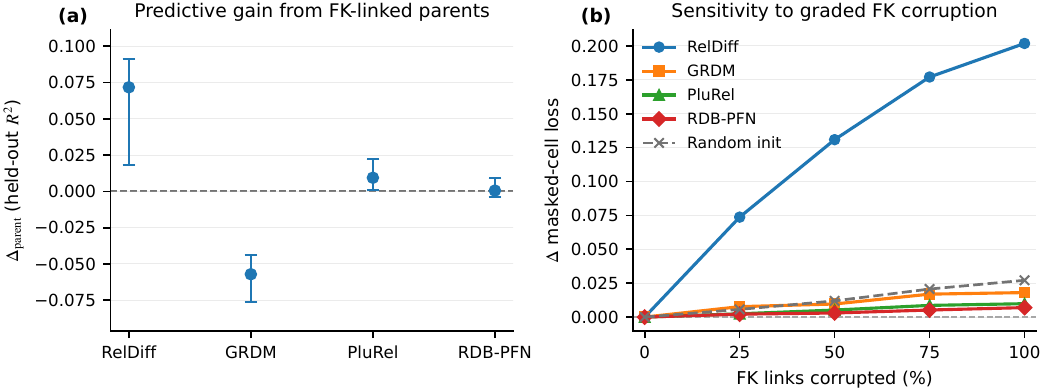}
    \caption{
    \textbf{From data-side relational signal to model-side structural reliance.}
    \textbf{Left:} median gain in held-out $R^2$ obtained by adding FK-linked parent
    attributes to a row-only predictor; error bars are 95\% bootstrap intervals.
    \textbf{Right:} increase in masked-cell reconstruction loss as an increasing fraction
    of \texttt{rel-stack} FK assignments is corrupted. RelDiff exhibits both the strongest
    parent-linked predictive signal and the steepest functional response to corruption.
    }
    \label{fig:data-mechanism}
\end{figure}

A binary intervention alone could still reflect an extreme out-of-distribution perturbation. We therefore vary the corrupted fraction $\rho\in\{0,0.25,0.50,0.75,1\}$ while holding the evaluation batches fixed.
Figure~\ref{fig:data-mechanism}(b) shows that RelDiff's loss increases monotonically with corruption strength. Linear fits give a slope $11$--$30\times$ larger than the other trained checkpoints. The random-init model and the three weakly reliant trained models remain comparatively flat.

Figure~\ref{fig:data-mechanism} places the data-side and model-side measurements next to one another. The corpus with the largest gain from linked-parent information is also the checkpoint whose reconstruction loss is most sensitive to progressively destroying FK structure.

\subsubsection{Reliance Across Real Databases}
\label{sec:reliance-cross-db}

The preceding experiments deliberately use one fixed \texttt{rel-stack} protocol so that binary corruption, graded corruption, pathway ablation, and the random-initialization control are directly comparable. We additionally ask whether the qualitative checkpoint ordering survives across real databases.

We therefore evaluate the original structural-intervention protocol over 27 fixed probe batches drawn from \texttt{rel-stack}, \texttt{rel-arxiv}, and \texttt{rel-ratebeer}. Because this is a different probe suite, its absolute loss values should not be compared directly with Table~\ref{tab:reliance}.

\begin{table}[t]
\caption{
\textbf{Structural reliance averaged over three real RelBench databases.}
Entries are mean original masked-cell loss and absolute changes under each intervention.
The experiment uses a broader 27-batch probe suite than Table~\ref{tab:reliance}.
}
\label{tab:reliance-real3}
\centering
\small
\begin{tabular}{lrrrr}
\toprule
Model &
Original &
$\Delta\mathcal{L}_{\mathrm{nbr}}$ &
$\Delta\mathcal{L}_{\mathrm{shuffle}}$ &
$\Delta\mathcal{L}_{\mathrm{remove}}$ \\
\midrule
RelDiff  & 0.339 & \textbf{+0.029} & \textbf{+0.069} & \textbf{+0.065} \\
GRDM     & 0.350 & +0.013 & $-0.005$ & $-0.007$ \\
PluRel   & 0.339 & +0.007 & +0.009 & +0.014 \\
RDB-PFN  & 0.351 & $-0.002$ & +0.003 & +0.007 \\
\bottomrule
\end{tabular}
\end{table}

The broader evaluation reproduces the aggregate separation: RelDiff is the only checkpoint for which FK corruption and internal neighbor ablation both produce substantial mean degradation. Importantly, however, the sign is not uniform across databases. For RelDiff, full FK shuffling changes reconstruction loss by $+0.165$ on \texttt{rel-stack}, $+0.067$ on \texttt{rel-ratebeer}, and $-0.112$ on \texttt{rel-arxiv}. Thus, the learned model is sensitive to relational structure across databases, but that structure is not always beneficial. On \texttt{rel-arxiv}, corrupting the learned relational prior improves reconstruction.

This distinction will matter again downstream: relational reliance should be interpreted as a learned inductive bias, not as a guarantee that the bias is correct for every target database.

%%%%%%%%%%%%%%%%%%%%%%%%%%%%%%%%%%%%%%%%%%%%%%%%%%%%%%%%%%%%%%%%%%%%%%%%%%%%%%%
% 3.3 MECHANISTIC LOCALIZATION
%%%%%%%%%%%%%%%%%%%%%%%%%%%%%%%%%%%%%%%%%%%%%%%%%%%%%%%%%%%%%%%%%%%%%%%%%%%%%%%

\subsection{Mechanistic Localization of Cross-Table Computation}
\label{sec:mechanism}

Structural corruption establishes dependence on relational information but perturbs every operation that consumes FK indices. We next intervene directly on RT's internal cross-table pathways to determine how the relational computation is organized.

\subsubsection{Pathway-Specific Interventions}
\label{sec:pathway-interventions}

The pathway experiment uses 12 fixed \texttt{rel-stack} batches, corresponding to 192 masked cells at context length 1,024. We compare four conditions:

\begin{enumerate}
    \item \textbf{Original}: both cross-table pathways operate normally;
    \item \textbf{No child aggregation}: the \texttt{nbr} output is zeroed;
    \item \textbf{No parent access}: the \texttt{feat} attention mask is restricted
    to same-row cells;
    \item \textbf{Both}: child aggregation and parent access are disabled simultaneously.
\end{enumerate}

These interventions distinguish two possible organizations. If \texttt{nbr} and \texttt{feat} provide independent routes by which relational information reaches the target, ablating both should be more damaging than ablating either route alone. If they instead form a serial computation, disabling a downstream route should make removal of
the upstream route redundant.

\subsubsection{A Serial Relational Pathway}
\label{sec:serial-path}

\begin{figure}[t]
    \centering
    \tikzset{
      mainnode/.style={
        rectangle, rounded corners=2pt, draw=black, thin, fill=white,
        text width=2.3cm, align=center, minimum height=1.0cm,
        font=\small, inner sep=2pt
      },
      annobox/.style={
        rectangle, draw=gray!70, thin, dashed, fill=gray!8,
        text width=2.5cm, align=center, font=\scriptsize, inner sep=2pt
      },
      arrlabel/.style={
        align=center, font=\scriptsize, text width=2.5cm
      },
      mainarrow/.style={
        -{Latex[length=1.8mm,width=1.2mm]}, thin, draw=black
      },
      summarybox/.style={
        rectangle, draw=black, thin, fill=gray!5, rounded corners=1pt,
        align=center, font=\scriptsize, inner sep=5pt, text width=8.6cm
      }
    }

    \begin{tikzpicture}[node distance=2.6cm, >=Latex]

    \node[mainnode] (child) {Linked child rows};
    \node[mainnode, right=of child] (parent) {Parent\\representation};
    \node[mainnode, right=of parent] (target) {Masked target cell};

    \draw[mainarrow] (child) -- (parent);
    \draw[mainarrow] (parent) -- (target);

    \coordinate (mid1) at ($(child)!0.5!(parent)$);
    \coordinate (mid2) at ($(parent)!0.5!(target)$);

    \node[arrlabel, above=0.25cm of mid1] (lab1)
      {\texttt{nbr}: child aggregation};
    \node[arrlabel, above=0.25cm of mid2] (lab2)
      {\texttt{feat}: parent access};

    \node[annobox, below=1.1cm of mid1] (anno1)
      {No child aggregation:\\
       $\mathcal{L}=0.386$\\
       $(\Delta\mathcal{L}=+0.128)$};

    \node[annobox, below=1.1cm of mid2] (anno2)
      {No parent access:\\
       $\mathcal{L}=0.453$\\
       $(\Delta\mathcal{L}=+0.195)$};

    \draw[gray!70, thin] (mid1) -- (anno1.north);
    \draw[gray!70, thin] (mid2) -- (anno2.north);

    \node[summarybox, below=0.9cm of anno1, xshift=2.3cm] (summary)
      {Original: $\mathcal{L}=0.258$
       \quad $\vert$ \quad
       Both ablated: $\mathcal{L}=0.453$
       $(\Delta\mathcal{L}=+0.195)$};

    \node[align=center, font=\small, below=0.35cm of summary] (concl1)
      {\textbf{Both ablated = no parent access}};
    \node[align=center, font=\small, below=0.12cm of concl1] (concl2)
      {$\Rightarrow$ serial, not independent};

    \end{tikzpicture}

    \caption{
    \textbf{Pathway-specific intervention localizes a serial cross-table computation.}
    Neighbor attention aggregates linked children into a parent representation, which
    subsequently reaches the target through parent access in \texttt{feat}. Blocking
    parent access makes additional \texttt{nbr} ablation behaviorally redundant.
    }
    \label{fig:pathway}
\end{figure}
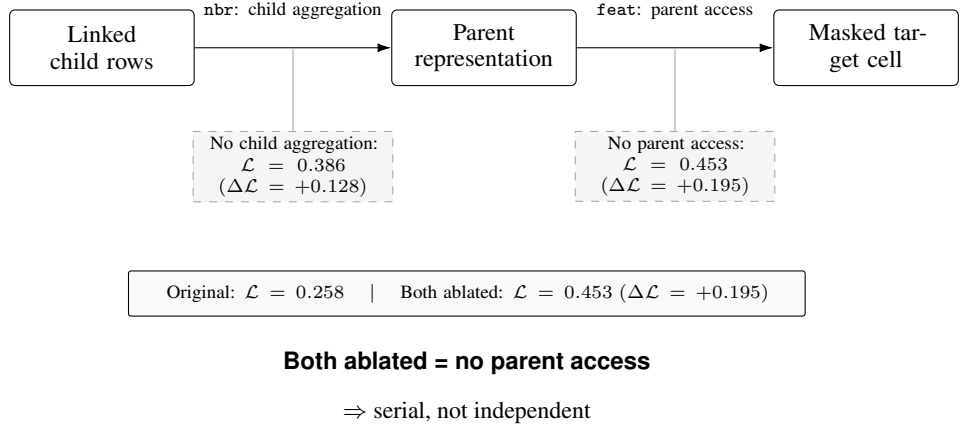

Figure~\ref{fig:pathway} supports the serial interpretation. RelDiff's original loss is $0.258$. Removing child aggregation raises it to $0.386$ ($\Delta\mathcal{L}=+0.128$), while removing parent access raises it to $0.453$
($+0.195$). Crucially, disabling \emph{both} routes also yields $0.453$, identical to parent-access ablation alone to six decimal places.

The equality is not caused by accidentally silencing \texttt{nbr}. Its activation magnitude remains essentially unchanged when parent access is blocked. The neighbor computation is therefore still performed, but its output can no longer influence the masked target. The functional pathway supported by these interventions is
\[
\text{linked child rows}
\;\xrightarrow{\;\texttt{nbr}\;}
\text{parent representation}
\;\xrightarrow{\;\texttt{feat}\;}
\text{masked target cell}.
\]

Thus, \texttt{nbr} should not be interpreted as an independent direct route from children to the target. It first enriches the representation of the linked parent, and \texttt{feat} subsequently exposes that representation to the queried cell.

\subsubsection{Fine-Grained Functional Specialization Across Heads}
\label{sec:head-specialization}

The pathway intervention establishes that the \texttt{nbr} stream is functionally important for RelDiff, but does not reveal whether this dependence is spread uniformly across its attention heads. We therefore ablate the eight \texttt{nbr} heads individually and measure the resulting change in reconstruction loss.

\begin{figure}[t]
    \centering
    \includegraphics[width=0.92\linewidth]{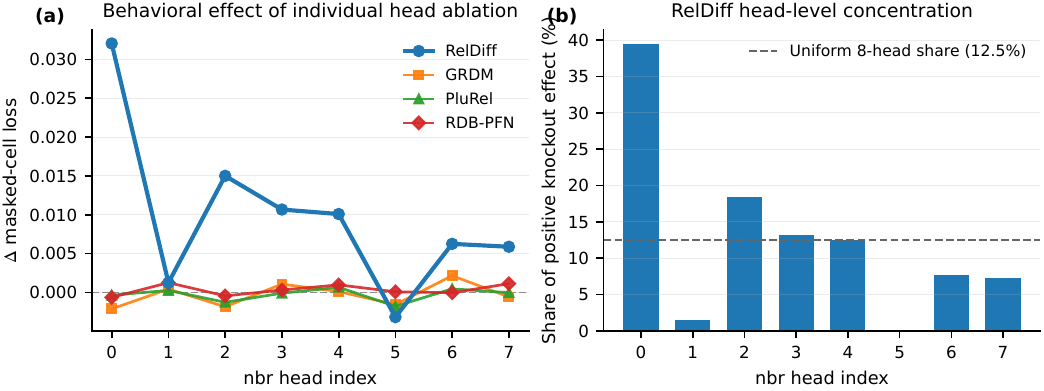}
    \caption{
    \textbf{Behavioral importance of individual \texttt{nbr} attention heads.}
    Individual head ablations reveal concentrated functional specialization in RelDiff.
    Its most important head accounts for approximately $39.5\%$ of the positive
    head-attributable loss increase, compared with $12.5\%$ under uniform contribution
    from eight heads.
    }
    \label{fig:head-knockout}
\end{figure}

Figure~\ref{fig:head-knockout} shows a clear concentration of functional importance for RelDiff. Ablating head 0 increases masked-cell loss by approximately $0.032$, while most individual heads in the other checkpoints have effects close to zero. Considering the positive head-attributable degradation in RelDiff, head 0 accounts for approximately $39.5\%$ of the total, compared with $12.5\%$ under a uniform eight-head contribution.

This concentration is specifically behavioral. Per-head weight-change Gini coefficients are nearly identical across checkpoints ($0.016$--$0.020$), and therefore do not recover the same specialization. This provides a useful methodological control: parameter movement can nominate components for investigation, but functional specialization must be established by intervention.

Additional weight-space, representation-similarity, and attention-distribution diagnostics are reported in Appendix~\ref{app:diagnostics}. We use them as complementary descriptions of checkpoint differences rather than as evidence of causal functional importance.

%%%%%%%%%%%%%%%%%%%%%%%%%%%%%%%%%%%%%%%%%%%%%%%%%%%%%%%%%%%%%%%%%%%%%%%%%%%%%%%
% 3.4 DOWNSTREAM
%%%%%%%%%%%%%%%%%%%%%%%%%%%%%%%%%%%%%%%%%%%%%%%%%%%%%%%%%%%%%%%%%%%%%%%%%%%%%%%

\subsection{Causal Link to Downstream Performance}
\label{sec:downstream}

The preceding experiments connect a property of the synthetic corpora to a learned cross-table computation. We now ask whether this computation explains downstream predictive behavior.

\subsubsection{Downstream Evaluation and Corrected Context Scaling}
\label{sec:benchmark}

We evaluate the four checkpoints using RT's in-context inference protocol on nine RelBench tasks across context lengths $\{100,200,512,1024,30000\}$. Six tasks are classification problems and are summarized
by AUROC.

At context length 1,024, mean classification AUROC is $0.748$ for RelDiff, $0.740$ for RDB-PFN, $0.723$ for PluRel, and $0.666$ for GRDM. Our original 30k-context evaluation, however, used only $n=256$ examples on two rare-label \texttt{rel-stack} tasks. This produced only six positive examples for \texttt{user-engagement} and ten for \texttt{user-badge}, making their AUROC estimates unstable.

We therefore repeated those evaluations with $n=2048$, yielding approximately 60 positive examples per task. The correction materially changes the context-scaling conclusion. The original RelDiff estimates decrease from $0.929$ to $0.865$ on \texttt{user-engagement} and from $0.847$ to $0.812$ on \texttt{user-badge};
on \texttt{user-engagement}, PluRel increases from $0.564$ to $0.652$ and RDB-PFN from $0.538$ to $0.642$. Already well-powered cells move by at most $0.007$.

After correction, \emph{all four} checkpoints decrease in mean classification AUROC from context 1,024 to 30k:
\[
\Delta\mathrm{AUC}_{30k-1k}
=
\begin{cases}
-0.011, & \text{RelDiff},\\
-0.038, & \text{GRDM},\\
-0.043, & \text{PluRel},\\
-0.069, & \text{RDB-PFN}.
\end{cases}
\]
We therefore do not claim that RelDiff improves indefinitely with increasing context. The supported result is narrower: among the four checkpoints, it is the most robust to the increase from 1,024 to 30k.

\begin{figure}[t]
    \centering
    \includegraphics[width=0.82\linewidth]{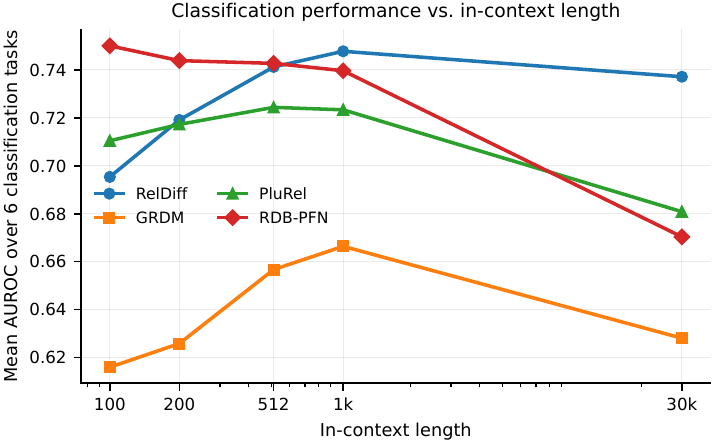}
    \caption{
    \textbf{Mean classification AUROC as a function of in-context length.}
    Each point averages the six classification tasks. The 30k points incorporate the
    $n=2048$ re-evaluation of the two rare-label \texttt{rel-stack} tasks. Under the
    corrected evaluation all four checkpoints degrade from context 1,024 to 30k, with
    RelDiff showing the smallest decline.
    }
    \label{fig:context-scaling}
\end{figure}

\subsubsection{Intervening on the Learned Mechanism}
\label{sec:downstream-intervention}

A benchmark association alone cannot establish that RelDiff's downstream behavior is caused by the relational computation identified above. We therefore rerun the downstream examples under the same two mechanistically informative interventions used during reconstruction:

\begin{itemize}
    \item \textbf{FK shuffled}: relational correspondences are corrupted while cell
    values are unchanged;
    \item \textbf{\texttt{nbr} ablated}: the input database is unchanged, but RT's
    internal child-aggregation output is zeroed.
\end{itemize}

For AUROC we report
\[
\Delta_{\mathrm{FK}}^{\mathrm{AUC}}
=
\mathrm{AUC}_{\mathrm{shuffle}}
-
\mathrm{AUC}_{\mathrm{original}},
\]
so a negative value indicates that correct relational structure contributes positively to prediction.

Table~\ref{tab:downstream-full} reports \texttt{rel-stack} user-engagement at both context 1,024 and 30k, using the same examples across intervention conditions.

\begin{table}[t]
\caption{
\textbf{Downstream intervention on \texttt{rel-stack} user-engagement.}
Entries are AUROC on identical examples under standard inference, FK shuffling, and
internal \texttt{nbr} ablation. The 30k evaluation uses $n=2048$ examples
(60 positives). $\Delta_{\mathrm{FK}}$ and $\Delta_{\mathrm{nbr}}$ are changes
relative to original inference.
}
\label{tab:downstream-full}
\centering
\small
\setlength{\tabcolsep}{4pt}
\begin{tabular}{rllrrrr}
\toprule
Context &
Model &
Original &
FK shuffled &
Nbr ablated &
$\Delta_{\mathrm{FK}}$ &
$\Delta_{\mathrm{nbr}}$ \\
\midrule
30k & RelDiff
& 0.865 & 0.579 & 0.624 & $-0.286$ & $-0.241$ \\
30k & GRDM
& 0.520 & 0.622 & 0.537 & +0.102 & +0.018 \\
30k & PluRel
& 0.652 & 0.650 & 0.636 & $-0.003$ & $-0.017$ \\
30k & RDB-PFN
& 0.642 & 0.643 & 0.634 & +0.001 & $-0.008$ \\
\midrule
1,024 & RelDiff
& 0.847 & 0.718 & 0.728 & $-0.129$ & $-0.119$ \\
1,024 & GRDM
& 0.609 & 0.618 & 0.622 & +0.009 & +0.013 \\
1,024 & PluRel
& 0.708 & 0.701 & 0.697 & $-0.007$ & $-0.011$ \\
1,024 & RDB-PFN
& 0.691 & 0.694 & 0.687 & +0.003 & $-0.004$ \\
\bottomrule
\end{tabular}
\end{table}

The effect on RelDiff is large at both context lengths. At 1,024, FK shuffling reduces AUROC from $0.847$ to $0.718$ ($-0.129$), and \texttt{nbr} ablation reduces it to $0.728$ ($-0.119$). At 30k, the corresponding effects grow to $-0.286$ and $-0.241$. FK corruption therefore removes RelDiff's intact advantage and pushes its
performance below the untouched PluRel and RDB-PFN checkpoints.

The comparison models behave differently. PluRel and RDB-PFN remain nearly invariant to FK shuffling at both context lengths. Their largest FK-shuffle change in the table is $0.007$ AUROC in magnitude. This mirrors the reconstruction experiment, in which the same checkpoints showed little functional dependence on FK structure.

The fact that the separation is already present at context 1,024 also rules out a simpler explanation in which the 30k intervention result arises only from the unusually large context window. Increasing context amplifies RelDiff's dependence on the relational mechanism, but does not create it.

Taken together, these interventions establish the strongest causal link in our analysis: holding the trained checkpoint and non-relational input values fixed, destroying relational correspondence or disabling the identified internal computation changes the downstream behavior of the model that learned strong relational reliance, while having little effect on the checkpoints that did not.

\subsubsection{When Relational Priors Hurt}
\label{sec:harmful-priors}

Relational reliance is not uniformly beneficial. On \texttt{rel-ratebeer} user-churn at context 1,024, FK shuffling improves RelDiff from $0.612$ to $0.673$ ($+0.061$ AUROC) and GRDM from $0.627$ to $0.691$ ($+0.064$), while PluRel and RDB-PFN change by at most $0.007$.

The same phenomenon appears in Table~\ref{tab:downstream-full}: on \texttt{rel-stack} user-engagement at 30k, shuffling FK structure improves GRDM from $0.520$ to $0.622$. These sign reversals are useful controls because they show that the intervention is not merely removing information that must always help. A learned relational
computation acts as an inductive bias. When the learned relation between topology and attributes matches the downstream problem, preserving relational structure can be beneficial; when it is mismatched, destroying that structure can improve prediction.

This behavior parallels the masked-reconstruction probe, where RelDiff is harmed by FK corruption on \texttt{rel-stack} and \texttt{rel-ratebeer} but improves on \texttt{rel-arxiv}. Relational reliance and relational usefulness are therefore distinct: a checkpoint may strongly use relational structure even when the particular structural prior it applies is poorly matched to a target database.

%%%%%%%%%%%%%%%%%%%%%%%%%%%%%%%%%%%%%%%%%%%%%%%%%%%%%%%%%%%%%%%%%%%%%%%%%%%%%%%
% 4. DISCUSSION + LIMITATIONS + CONCLUSION
%%%%%%%%%%%%%%%%%%%%%%%%%%%%%%%%%%%%%%%%%%%%%%%%%%%%%%%%%%%%%%%%%%%%%%%%%%%%%%%

\section{Discussion and Limitations}
\label{sec:discussion}

Our results distinguish \emph{architectural access to relational information} from \emph{learned functional reliance on relations}. All four checkpoints use the same FK-aware RT architecture and therefore have access to the same cross-table computational routes, yet they respond very differently when relational structure is corrupted or internally ablated. The checkpoint trained on the corpus with the strongest measurable parent-linked predictive signal also develops the strongest dependence on correct FK structure. That dependence can be localized to a concrete serial cross-table pathway, and intervention on the same computation substantially changes downstream behavior. The strongest causal evidence in our study therefore concerns the mechanism-to-behavior link: when the trained checkpoint is held fixed, disrupting the relational structure or the corresponding internal computation changes the prediction.

The data-side analysis suggests a broader design principle for synthetic relational pretraining. Structural complexity alone is not sufficient. A generated corpus may contain many tables, foreign keys, diverse schemas, or heterogeneous cardinalities while still allowing the pretraining objective to be solved largely from row-local regularities. To create optimization pressure for cross-table computation, relational context must provide
predictive information that is not already available locally. Our diversity analysis supports this distinction: corpus diversity does not reproduce the ordering of functional relational reliance across checkpoints (Appendix~\ref{app:diversity}). This suggests that future relational data generators should be evaluated not only by how much structural variation they produce, but also by whether that structure carries predictive information
relevant to the model's pretraining objective.

The results also emphasize that relational reliance and relational usefulness are not equivalent. On some databases and tasks, corrupting relational structure improves reconstruction or downstream accuracy. The learned relational computation should therefore be viewed as an inductive bias rather than an intrinsically beneficial capability. When the statistical dependencies learned during pretraining align with the target database,
the bias can improve prediction; when they are mismatched, the same mechanism can be harmful. This distinction is important for evaluating relational foundation models: sensitivity to structure demonstrates that the model is using relations, but does not by itself establish that the learned relational prior transfers correctly to every schema or task.

Our study has several limitations. First, we analyze a single relational architecture and a single masked-cell reconstruction objective. The specific \texttt{nbr}$\rightarrow$\texttt{feat} computation identified here is therefore RT-specific, although the data-side notion of relational predictive necessity is not tied to a particular architecture and may generalize to other relational pretraining settings. Second, each generator is represented by one trained checkpoint. We can therefore measure paired intervention effects within each checkpoint, but we cannot estimate variability across training seeds, independently generated corpora, or repeated pretraining runs. Third, the four synthetic corpora differ along multiple dimensions simultaneously. The association between corpus predictive structure and learned relational reliance is therefore a controlled observational attribution rather than a randomized intervention on the training distribution. A stronger causal test would begin from the same generated corpus, selectively destroy FK-conditioned predictive coupling while preserving marginal and row-local statistics, retrain from the same initialization and budget, and test whether relational reliance disappears. The complementary sufficiency test would inject such coupling into a weakly relational corpus and ask whether reliance emerges. Finally, the available GRDM corpus contains the semantic-type preprocessing confound described in Section~\ref{sec:evaluation-setting}. We therefore retain GRDM as an informative comparison but avoid interpreting its absolute ranking as an isolated estimate of the quality of its relational generation procedure.

\paragraph{Conclusion.}
Synthetic-data benchmarks can reveal which pretraining corpus produces the strongest model, but not what properties of that corpus produced a capability. We trace one such capability across four levels of analysis: relational predictive signal is measurable directly in the synthetic pretraining distribution; the corresponding checkpoint develops load-bearing dependence on FK structure; that dependence can be localized to a specific cross-table computation; and intervening on the same mechanism changes downstream behavior. More broadly, our results motivate attributing the effects of synthetic pretraining not only to generator identity, but to measurable properties of the generated distribution that create optimization pressure for particular computations to be learned.

\bibliographystyle{unsrt}
\bibliography{references}

\newpage
\appendix
\section{Generator and Corpus Details}
\label{app:generator-details}

The main text deliberately treats generator identity as a source of different pretraining distributions rather than as the explanatory variable itself. Here we provide additional details on the four generators and document the semantic-type issue affecting the available GRDM corpus.

\subsection{Generator Characteristics}
\label{app:generator-characteristics}

The four generators differ in both how relational structure is obtained and how attribute reference-conditioned: generation is tied to structural properties of an existing relational database. PluRel and RDB-PFN instead construct relational databases from synthetic priors without requiring a reference database.

Table~\ref{tab:generator-summary} summarizes the distinctions relevant to this study. These categories are descriptive rather than causal: each generator changes multiple properties of the resulting pretraining distribution simultaneously.

\begin{table}[ht]
\caption{
\textbf{Summary of the four synthetic pretraining generators.}
``Reference-conditioned'' denotes generation tied to the schema or relational structure
of an existing database; ``reference-free'' denotes generation from a synthetic prior.
}
\label{tab:generator-summary}
\centering
\small
\begin{tabular}{llll}
\toprule
Generator & Reference data & Relational structure & Value generation \\
\midrule
RelDiff
& Yes
& fitted to reference FK graph
& joint graph-conditioned diffusion \\

GRDM
& Yes
& reference structure / degree statistics
& graph-conditional diffusion \\

PluRel
& No
& procedural relational sampler
& conditional causal mechanisms \\

RDB-PFN
& No
& hand-designed relational prior
& SCM-style synthetic prior \\
\bottomrule
\end{tabular}
\end{table}

\subsection{GRDM Semantic-Type Conversion}
\label{app:grdm-types}

RT contains type-specific input pathways for numerical, textual, datetime, and boolean values, including a dedicated MiniLM-based text encoder. We therefore audited the semantic types present in the exported pretraining corpora. Table~\ref{tab:type-census} reports the resulting census.

\begin{table}[ht]
\caption{
\textbf{Column-type census in the available synthetic corpora.}
Counts are over the sampled databases used in the audit. The available GRDM export
contains no preserved text/categorical columns: string and datetime fields are represented
numerically.
}
\label{tab:type-census}
\centering
\small
\begin{tabular}{lrrr}
\toprule
Corpus & Numerical & Text / categorical & Boolean \\
\midrule
RelDiff  & 512  & 234 & 0 \\
GRDM     & 518  & 0   & 0 \\
PluRel   & 1132 & 0   & 518 \\
RDB-PFN  & 1081 & 0   & 0 \\
\bottomrule
\end{tabular}
\end{table}

This difference is relevant because the GRDM checkpoint received no training signal through RT's text pathway. In particular, reference databases for which the RelDiff corpus retains string-valued attributes have no corresponding preserved string columns in the available GRDM export.

Consequently, comparisons involving GRDM combine differences in relational generation with differences in semantic-type exposure. We therefore retain GRDM as an informative comparison but do not interpret its absolute ranking as an isolated effect of relational coupling or generator design.

%%%%%%%%%%%%%%%%%%%%%%%%%%%%%%%%%%%%%%%%%%%%%%%%%%%%%%%%%%%%%%%%%%%%%%%%%%%%%%%%%%%%%%%%%%%%%%%%%%%%%%%%%%%%%%%%

\section{Additional Corpus-Level Analyses}
\label{app:corpus-analyses}

\subsection{Task-Level Predictive-Necessity Distributions}
\label{app:necessity-distribution}

\begin{figure}[ht]
    \centering
    \includegraphics[width=0.92\linewidth]{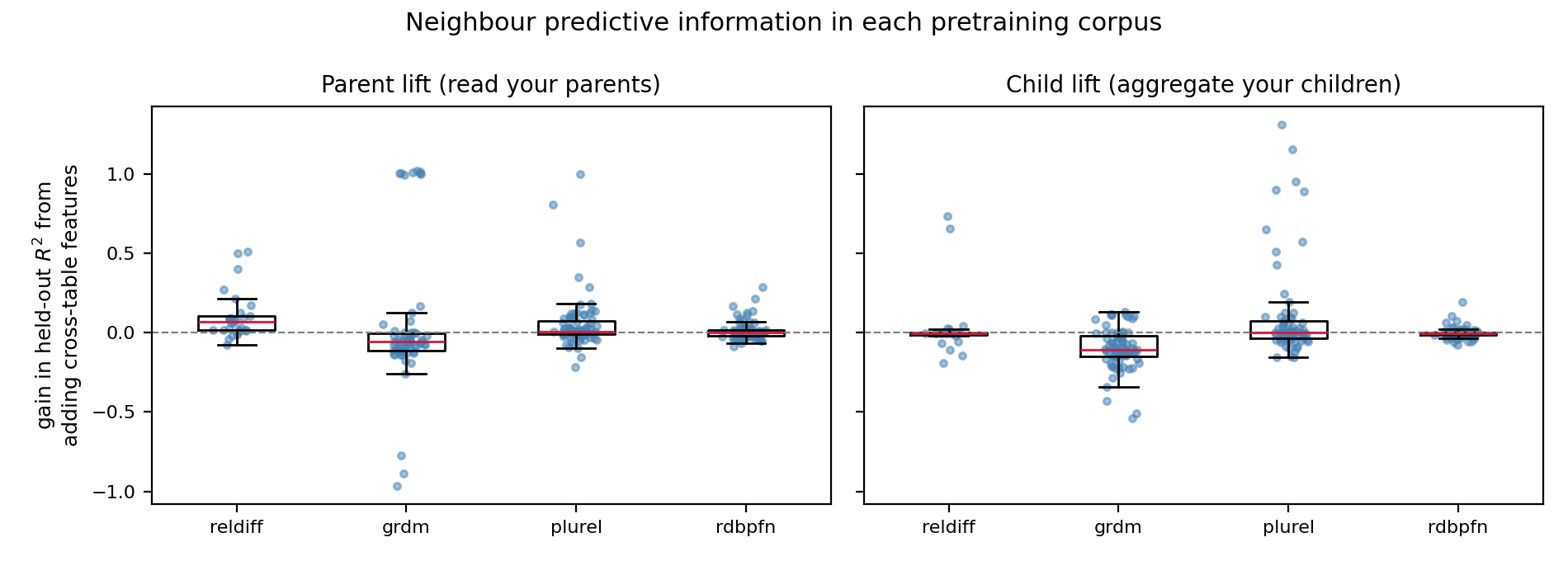}
    \caption{
    \textbf{Task-level relational predictive gains.}
    Each point corresponds to one target-column prediction problem. RelDiff exhibits
    substantially larger gains from linked-parent attributes, whereas the simple child
    aggregates considered here provide little additional predictive signal.
    }
    \label{fig:necessity-full}
\end{figure}

Table~\ref{tab:necessity} summarizes relational predictive necessity using medians across target-column prediction problems. Figure~\ref{fig:necessity-full} shows the underlying task-level values.

Each point corresponds to one held-out target-column problem and measures the change in $R^2$ obtained by augmenting row-only features with relational information. The task-level distribution is important because an aggregate median alone could in principle be driven by a small number of unusual generated databases.

The distribution supports the aggregate result. RelDiff exhibits more frequent and larger positive gains from linked-parent attributes, whereas RDB-PFN is strongly concentrated near zero. Child-aggregate gains remain close to zero across the corpora. Thus, the data-side result is not simply a single extreme target-column problem.

\subsection{Corpus Diversity as an Alternative Explanation}
\label{app:diversity}

Relational predictive necessity is not the only corpus-level property that could explain checkpoint differences. A natural alternative is that RelDiff simply generates a more diverse training distribution. We therefore measure feature-, table-, and cross-table diversity directly in the generated databases.

The diversity metrics are based on mean pairwise embedding distances and do not use representations from the trained RT checkpoints.

\begin{table}[ht]
\caption{
\textbf{Diversity of the generated corpora.}
Values are mean pairwise embedding distances along feature-, table-, and cross-table
axes. RelDiff has the largest measured diversity, but the ordering of the remaining
corpora does not reproduce the ordering of functional relational reliance.
}
\label{tab:diversity}
\centering
\small
\begin{tabular}{lccc}
\toprule
Corpus & Feature & Table & Cross-table \\
\midrule
RelDiff & \textbf{1.10} & \textbf{0.57} & \textbf{1.04} \\
GRDM    & 0.69 & 0.43 & 0.71 \\
PluRel  & 0.79 & 0.34 & 0.85 \\
RDB-PFN & 0.82 & 0.16 & 0.81 \\
\bottomrule
\end{tabular}
\end{table}

RelDiff is the most diverse corpus under all three measurements. Diversity alone, however, does not explain the functional results. Most notably, PluRel has greater cross-table diversity than GRDM ($0.85$ versus $0.71$), despite exhibiting little dependence on FK structure under the interventions in Sec.~\ref{sec:reliance}.
RDB-PFN similarly has greater feature diversity than PluRel while remaining nearly insensitive to structural corruption.

Diversity and predictive necessity therefore capture different properties. Diversity asks how broadly the generated data vary; predictive necessity asks whether information is coupled across relations in a way that improves prediction of the pretraining target.

\begin{figure}[ht]
    \centering
    \includegraphics[width=0.78\linewidth]{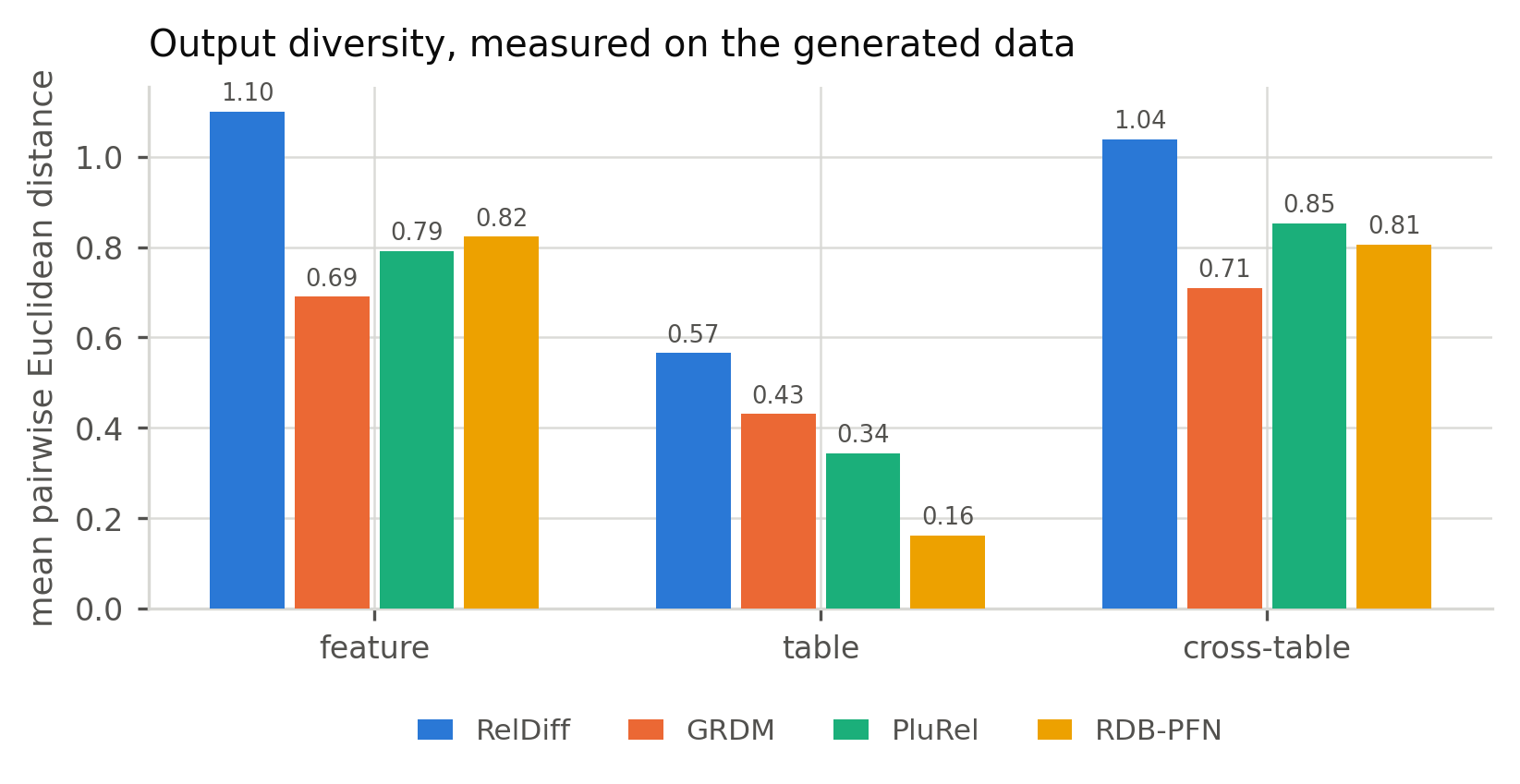}
    \caption{
    \textbf{Output diversity measured directly in the generated databases.}
    RelDiff has the largest diversity on all three axes, but diversity alone does not
    reproduce the ordering of functional relational reliance.
    }
    \label{fig:corpus-diversity}
\end{figure}

%%%%%%%%%%%%%%%%%%%%%%%%%%%%%%%%%%%%%%%%%%%%%%%%%%%%%%%%%%%%%%%%%%%%%%%%%%%%%%%%%%%%%%%%%%%%%%

\section{Additional Checkpoint Diagnostics}
\label{app:diagnostics}

The main mechanistic claims are based on behavioral interventions. This section reports complementary diagnostics of parameter movement, representation similarity, and attention distributions. These measurements help characterize where checkpoints differ, but they do not by themselves establish that a component is functionally necessary.

\subsection{Weight-Space Specialization}
\label{app:weights}

Because all four checkpoints begin from the same initialization, parameter displacement provides a simple diagnostic of which components were altered most strongly during pretraining. Raw $L_2$ distances are not directly comparable across modules of different sizes, so we normalize by parameter count and report the RMS deviation
\[
D_{\mathrm{RMS}}(W)
=
\sqrt{
\frac{1}{|W|}
\left\|W-\bar{W}\right\|_2^2
},
\]
where $\bar{W}$ is the corresponding tensor in the four-checkpoint mean.

\begin{table}[ht]
\caption{
\textbf{Per-parameter checkpoint deviation by RT component.}
Entries are RMS deviations from the four-checkpoint mean, multiplied by $10^3$.
Among the attention streams, RelDiff has its largest deviation in \texttt{nbr},
whereas GRDM has its largest deviation in \texttt{feat}.
}
\label{tab:weight-rms}
\centering
\small
\begin{tabular}{lccccc}
\toprule
Model & Encoders & Col & Feat & Nbr & FFN \\
\midrule
GRDM    & 4.386 & 3.440 & \textbf{4.047} & 3.129 & 3.525 \\
PluRel  & 2.345 & \textbf{2.893} & 2.636 & 2.198 & 2.504 \\
RDB-PFN & 2.311 & \textbf{3.142} & 2.455 & 2.142 & 2.445 \\
RelDiff & 3.690 & 3.025 & 3.315 & \textbf{3.622} & 3.089 \\
\bottomrule
\end{tabular}
\end{table}

Normalizing by parameter count changes an interpretation suggested by raw $L_2$ norms: the encoders are not uniformly insensitive to the pretraining corpus. Their per-parameter movement is substantial for the reference-conditioned checkpoints.

The narrower observation relevant to the main mechanism nevertheless survives: among RT's attention streams, RelDiff's largest RMS displacement is in \texttt{nbr}, whereas GRDM's is in \texttt{feat}. We treat this only as a localization cue, not as evidence that those parameters are functionally important.

\begin{figure}[ht]
    \centering
    \includegraphics[width=\linewidth]{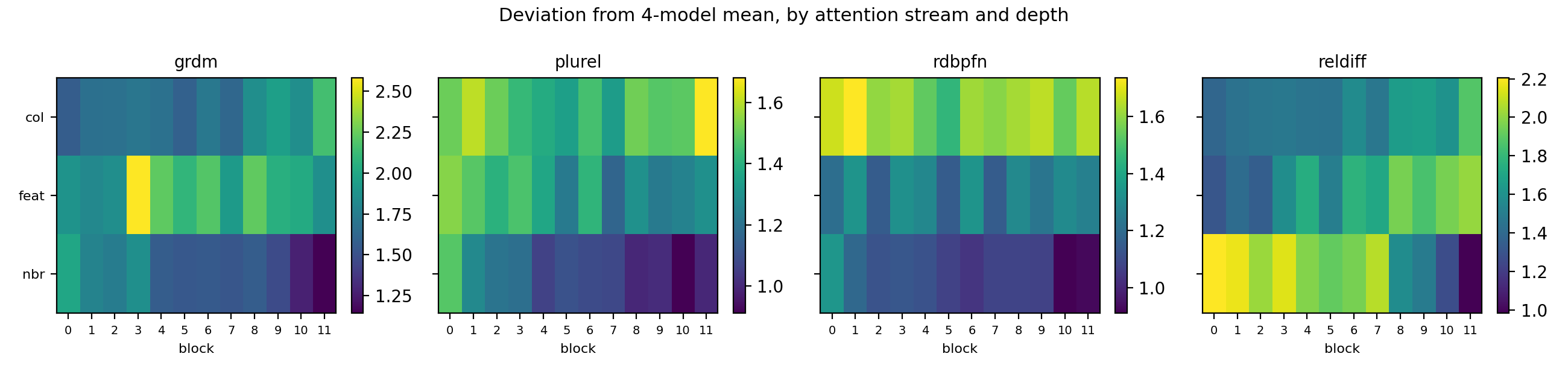}
    \caption{
    \textbf{Checkpoint deviation by attention stream and depth.}
    Weight-space differences identify components that were affected differently during
    pretraining. Functional importance is established separately by intervention.
    }
    \label{fig:weight-stream-depth}
\end{figure}

\subsection{Representation Similarity}
\label{app:cka}

Parameter similarity does not imply similarity of the computations performed on real inputs. We therefore compute linear centered-kernel alignment (CKA) between residual-stream activations from different checkpoints evaluated on identical inputs. Approximately 20,000 non-padding cell activations are collected for each comparison. We summarize blocks 0--5 as early layers and blocks 6--11 as late layers.

\begin{table}[ht]
\caption{
\textbf{Pairwise linear CKA on identical real-data inputs.}
Early and late values average blocks 0--5 and 6--11, respectively.
}
\label{tab:cka}
\centering
\small
\begin{tabular}{lcc}
\toprule
Checkpoint pair & Early CKA & Late CKA \\
\midrule
GRDM / RelDiff    & 0.944 & \textbf{0.824} \\
PluRel / RelDiff  & 0.944 & 0.810 \\
GRDM / PluRel     & 0.938 & 0.718 \\
PluRel / RDB-PFN  & 0.909 & 0.707 \\
RDB-PFN / RelDiff & 0.870 & 0.587 \\
GRDM / RDB-PFN    & 0.808 & 0.406 \\
\bottomrule
\end{tabular}
\end{table}

Checkpoint representations generally diverge with depth. More importantly, representation similarity and parameter-space similarity do not induce the same clustering. PluRel and RDB-PFN are comparatively close in parameter space, yet RDB-PFN is the largest late-layer functional outlier under CKA. Conversely, GRDM and RelDiff have the largest late-layer CKA.

This disagreement illustrates why parameter-space proximity is insufficient to infer similarity of learned computation.

\subsection{Attention Within the Permitted FK Neighborhood}
\label{app:attention-entropy}

RT's \texttt{nbr} attention is masked so that a head may attend only to structurally permitted FK-linked cells. Consequently, observing attention on relational edges is not itself evidence of learning: the support is imposed architecturally.

We instead ask how each checkpoint distributes probability \emph{within} its permitted neighborhood. We recompute attention probabilities from the \texttt{nbr} query/key projections and report normalized entropy, for which a value of one corresponds to uniform attention, together with the probability mass assigned to the highest-weight
neighbor.

\begin{table}[ht]
\caption{
\textbf{Within-neighborhood attention statistics.}
Lower normalized entropy indicates more selective attention among the FK-linked cells
permitted by the architectural mask.
}
\label{tab:attention-entropy}
\centering
\small
\begin{tabular}{lcc}
\toprule
Model & Normalized entropy & Top-1 mass \\
\midrule
RelDiff & \textbf{0.843} & \textbf{0.396} \\
GRDM    & 0.858 & 0.394 \\
RDB-PFN & 0.890 & 0.361 \\
PluRel  & 0.898 & 0.361 \\
\bottomrule
\end{tabular}
\end{table}

RelDiff and GRDM exhibit sharper within-neighborhood attention than PluRel and RDB-PFN. However, GRDM's relatively selective attention does not translate into comparable functional dependence under FK corruption. Attention concentration therefore characterizes how the permitted neighborhood is weighted, but is not itself evidence that the resulting relational computation is load-bearing.

%%%%%%%%%%%%%%%%%%%%%%%%%%%%%%%%%%%%%%%%%%%%%%%%%%%%%%%%%%%%%%%%%%%%%%%%ù

\section{Negative Result: Input-Decodable Linear Probes}
\label{app:probes}

We also investigated whether differences between pretraining corpora could be identified using linear probes on frozen RT representations. We considered semantic type, table identity, parent-table identity, and row degree, asking whether these quantities become more linearly decodable after pretraining under particular synthetic distributions.

These probes do not discriminate meaningfully between the checkpoints. All tested properties are already at or near ceiling in a randomly initialized RT, and improvements relative to the random-init baseline are below $0.014$; for two trained checkpoints the measured gain is negative.

The failure is structural rather than simply statistical. RT is a residual architecture, so information available in the original cell representation can remain linearly accessible at later layers even if the network has not learned a computation involving that information. Moreover, quantities such as parent-table identity and row degree can be derived directly from the FK index structure supplied as input.

Consequently, high probe accuracy for these targets does not demonstrate that RT has learned a relational computation. We therefore exclude these probes from the main mechanistic evidence. A more discriminative probe target would need to depend on information unavailable in the queried cell's input representation and require actual cross-table computation---for example, a property of the values contained in linked parent rows.

%%%%%%%%%%%%%%%%%%%%%%%%%%%%%%%%%%%%%%%%%%%%%%%%%%%%%%%%%%%%%%%%%%%%%%%%%%%%%%%%%%%%%%%%%%%%%%%%%%%%%%%%

\section{Database-Specific Sign of Structural Reliance}
\label{app:reliance-full}

Section~\ref{sec:reliance-cross-db} reports the three-database aggregate structural intervention. Here we emphasize the database-specific sign of the effect, which is important for distinguishing \emph{relational reliance} from \emph{beneficial relational reliance}.

For RelDiff, the change in masked-cell reconstruction loss under FK shuffling is
\[
\Delta\mathcal{L}_{\mathrm{shuffle}}
=
\begin{cases}
+0.165, & \texttt{rel-stack},\\
+0.067, & \texttt{rel-ratebeer},\\
-0.112, & \texttt{rel-arxiv}.
\end{cases}
\]
The corresponding absolute changes for the other checkpoints are at most $0.027$ in this probe suite.

A positive value means that correct FK structure helps reconstruction, whereas a negative value means that destroying the structure improves it. The \texttt{rel-arxiv} sign reversal therefore does not indicate absence of a relational effect. Rather, the checkpoint's predictions change substantially when relational correspondences are
destroyed, but in a beneficial direction.

This distinction motivates our terminology throughout the paper: \emph{reliance} measures whether a learned computation depends on relational structure, not whether that dependence necessarily improves performance.

%%%%%%%%%%%%%%%%%%%%%%%%%%%%%%%%%%%%%%%%%%%%%%%%%%%%%%%%%%%%%%%%%%%%%%%%%%%%%%%%%%%%%%%%%%%%%%%%%%ùù

\section{Additional Downstream Evaluation Details}
\label{app:downstream-full}

\subsection{Rare-Label Re-Evaluation at 30k Context}
\label{app:context-scaling}

The initial 30k-context evaluation used $n=256$ examples on two rare-label \texttt{rel-stack} classification tasks. This produced only six positive examples for \texttt{user-engagement} and ten for \texttt{user-badge}, making AUROC highly sensitive to individual predictions.

We therefore reran both tasks with $n=2048$. The corrected evaluation contains 60 positive examples for \texttt{user-engagement} and 64 for \texttt{user-badge}. The largest changes occur precisely in the previously underpowered cells:

\begin{itemize}
    \item RelDiff \texttt{user-engagement}: $0.929 \rightarrow 0.865$;
    \item RelDiff \texttt{user-badge}: $0.847 \rightarrow 0.812$;
    \item PluRel \texttt{user-engagement}: $0.564 \rightarrow 0.652$;
    \item RDB-PFN \texttt{user-engagement}: $0.538 \rightarrow 0.642$.
\end{itemize}

Evaluation cells that were already well powered change by at most $0.007$. This pattern is consistent with instability caused by the small number of positive examples rather than a systematic change to the evaluation protocol.

All context-scaling results in the main paper use these corrected values. In particular, we make no claim that RelDiff's mean classification performance improves from context 1,024 to 30k.

\subsection{Additional Example of a Harmful Relational Prior}
\label{app:harmful-priors}

Section~\ref{sec:harmful-priors} reports that disrupting relational structure can improve performance when the learned structural prior is mismatched to the downstream problem.

On \texttt{rel-ratebeer} user-churn at context 1,024, FK shuffling changes AUROC from $0.612$ to $0.673$ for RelDiff and from $0.627$ to $0.691$ for GRDM, corresponding to improvements of $+0.061$ and $+0.064$. PluRel and RDB-PFN change by at most $0.007$ under the same intervention.

This downstream sign reversal mirrors the reconstruction-level result in Appendix~\ref{app:reliance-full}. Together they illustrate that learned relational structure functions as an inductive bias: sensitivity to relations can produce either gains or losses depending on how well the learned prior matches the target database and task.

%%%%%%%%%%%%%%%%%%%%%%%%%%%%%%%%%%%%%%%%%%%%%%%%%%%%%%%%%%%%%%%%%%%%%%%%%%%%%%%%%%%%%%%%%%%%%%%%%%ù

\section{Implementation and Reproducibility Details}
\label{app:reproducibility}

This section records the evaluation protocols underlying the primary results. Within each intervention experiment, all checkpoints are evaluated on identical examples, masks, and randomly selected corruption positions so that changes are paired at the example level.

\paragraph{Predictive-necessity analysis.}
For each of the four synthetic corpora, we sample 40 generated databases. Each eligible target column defines a held-out prediction problem. Identical LightGBM procedures are used for the row-only, row-plus-parent, and row-plus-child-aggregate feature sets. Child features consist of mean, standard deviation, and count aggregates over linked rows. Held-out $R^2$ scores are clipped to $[-1,1]$ before relational gains are computed. Tables report medians across target-column tasks, and parent gains include 95\% bootstrap confidence intervals.

\paragraph{Structural interventions.}
All checkpoints receive identical preprocessed inputs. FK shuffling permutes FK-neighbor index assignments among the selected row positions while leaving cell values unchanged. FK removal sets the corresponding relational indices to the missing-link value. Internal neighbor ablation leaves the input database unchanged and sets the output of the \texttt{nbr} module to zero during inference.

\paragraph{Random-initialization and graded-corruption experiment.}
The controlled \texttt{rel-stack} experiment uses 24 fixed batches containing 384 masked cells at context length 1,024. The random-initialization control uses the same RT architecture without pretraining. Graded corruption evaluates $\rho\in\{0,0.25,0.50,0.75,1\}$ on exactly the same examples; $\rho$ denotes the fraction of FK assignments selected for corruption.

\paragraph{Pathway intervention.}
The pathway-specific experiment uses 12 fixed \texttt{rel-stack} batches containing 192 masked cells at context length 1,024. Child aggregation is disabled by zeroing the output of \texttt{nbr}. Parent access is disabled by modifying the \texttt{feat} attention mask so that only same-row cells remain accessible. The joint condition applies both interventions simultaneously.

\paragraph{Individual-head intervention.}
For the head-level experiment, each of the eight \texttt{nbr} attention heads is ablated individually while all other model components remain unchanged. Functional importance is measured by the resulting increase in masked-cell reconstruction loss relative to the same intact batches. When reporting the fraction of RelDiff's positive head-attributable effect, negative knockout changes are excluded from the denominator rather than treated
as negative importance.

\paragraph{Broader real-database probe.}
The cross-database structural-reliance experiment uses 27 fixed probe batches drawn from \texttt{rel-stack}, \texttt{rel-arxiv}, and \texttt{rel-ratebeer}. This probe suite is distinct from the 24-batch controlled \texttt{rel-stack} experiment; absolute loss values between the two protocols are therefore not directly compared.

\paragraph{Downstream evaluation.}
Downstream experiments use RT's standard in-context inference protocol on nine RelBench tasks and context lengths $100$, $200$, $512$, $1024$, and $30000$. Classification performance is measured by AUROC. The corrected 30k evaluations of the two rare-label \texttt{rel-stack} tasks use $n=2048$ examples. For every downstream intervention, the original, FK-shuffled, and \texttt{nbr}-ablated conditions are evaluated on identical
examples.

\paragraph{Scope of statistical comparisons.}
The predictive-necessity experiment reports bootstrap uncertainty across target-column prediction problems. The mechanistic intervention experiments instead use paired, fixed-batch comparisons: the same masked cells are evaluated before and after each intervention. Because each generator is represented by one pretrained checkpoint, these experiments quantify within-checkpoint intervention effects rather than variance across independent pretraining seeds.

\end{document}